\documentclass[11pt]{article}

\usepackage[margin=1in]{geometry}
\usepackage{graphicx}
\usepackage[utf8]{inputenc}
\usepackage[T1]{fontenc}
\usepackage{amsmath,amssymb}
\usepackage{booktabs}
\usepackage{natbib}
\usepackage{hyperref}
\usepackage{xcolor}
\usepackage{caption}
\usepackage{subcaption}
\usepackage{multirow}
\usepackage{enumitem}
\usepackage{url}
\usepackage[expansion=false]{microtype}
\usepackage{lineno}
\usepackage{authblk}

\hypersetup{
  colorlinks=true,
  linkcolor=blue!60!black,
  citecolor=blue!60!black,
  urlcolor=blue!60!black,
}

\title{\Large\bfseries The first global agricultural field boundary map at 10\,m resolution}

\author[1]{Caleb Robinson\thanks{Corresponding author: \texttt{caleb.robinson@microsoft.com}}}
\author[2]{Gedeon Muhawenayo}
\author[3]{Subash Khanal}
\author[4]{Zhanpei Fang}
\author[5]{Isaac Corley}
\author[2]{Ana M. Tárano}
\author[6]{Lyndon Estes}
\author[5]{Jennifer Marcus}
\author[3]{Nathan Jacobs}
\author[2,5]{Hannah Kerner}
\author[1]{Inbal Becker-Reshef}
\author[1]{Juan M. Lavista Ferres}

\affil[1]{Microsoft AI for Good Research Lab}
\affil[2]{Arizona State University}
\affil[3]{Washington University in St.\ Louis}
\affil[4]{Oregon State University}
\affil[5]{Taylor Geospatial}
\affil[6]{Clark University}

\date{}

\begin{document}
\maketitle
\thispagestyle{empty}

\begin{abstract}
\noindent
The agricultural field is the natural unit at which crops are planted, managed, regulated, and reported, yet most global remote-sensing products for agriculture are only available at the pixel level. While some high-quality field-level data products exist, they come from parcel registries covering only parts of Europe or from ML-derived products for individual countries.
No openly available, globally consistent map of agricultural field boundaries exists to date.
Here we present the first global field boundary dataset at 10\,m resolution for the years 2024 and 2025, comprising 3.17\,billion remote-sensing field polygons (1.62\,B in 2024 and 1.55\,B in 2025) across 241 countries and territories, produced by applying a U-Net segmentation model trained on the Fields of The World dataset to cloud-free Sentinel-2 mosaics.
Validated against ground-truth field boundaries in 24 countries, the map achieved a mean pixel-level recall of 0.85 with 14 countries exceeding 0.90. Evaluation against full-country ground-truth datasets in Austria, Latvia, and Finland yielded F1 scores of 0.89, 0.88, and 0.74, respectively.
Because reference data for global validation is inherently incomplete, we accompanied the map with a 500\,m confidence layer that identifies regions where predictions are reliable.
We release the dataset openly as three global maps: the confidence-thresholded default field boundary dataset, the full unfiltered dataset, and the continuous-valued confidence raster. These maps provide the first globally consistent field-level unit of analysis for crop monitoring, food security, and downstream agricultural science.
\end{abstract}

\section*{Main}

The agricultural field is the natural unit at which crops are planted, managed, harvested, traded, regulated, and reported.
Yet most global remote-sensing products for agriculture (cropland masks~\citep{potapov2022global}, vegetation indices~\citep{didan2015mod13q1}, crop condition monitoring~\citep{becker2019geoglam}) operate at the pixel level and treat the landscape as a continuous surface.
The result is a persistent mismatch between how agriculture is monitored from space and how it is actually organized on the ground.
A global, openly available field boundary map would supply that missing unit, but no such map exists to date.
Field-level spatial units enable crop type mapping, yield estimation, pest and disease surveillance, resource use tracking, and the measurement, reporting, and verification (MRV) of conservation and climate programs~\citep{nakalembe2023considerations}.
National statistics agencies depend on field boundary data for survey design, and multi-year boundary maps reveal socioeconomic dynamics such as farm consolidation and fragmentation~\citep{sullivan2023large, estes2022high}.
Regulatory frameworks, including the European Union Deforestation Regulation~\citep{eu_deforestation_regulation_2023}, increasingly require spatially explicit evidence of agricultural land use, raising the demand for globally consistent field boundary data.

While the global number of farms has been estimated at approximately 570\,million~\citep{lowder2016number} (most of them smallholder operations under 2\,ha~\citep{lesiv2019estimating}), the corresponding number of agricultural fields worldwide is unknown. This is a basic missing baseline for quantitative analysis of global food production, land use, and farm structure.
Existing data from government-sourced cadastral and Land Parcel Identification System (LPIS) records provide high-quality boundaries in parts of Europe, but remain unavailable, incomplete, or restricted in most of the world~\citep{kerner2025fields}.
Further, manual digitization of field boundaries from satellite imagery is slow, expensive, and must be repeated as land use changes over time, making it impractical at continental to global scales~\citep{estes2022high}.

\begin{figure*}[t!]
  \centering
  \includegraphics[width=1\linewidth]{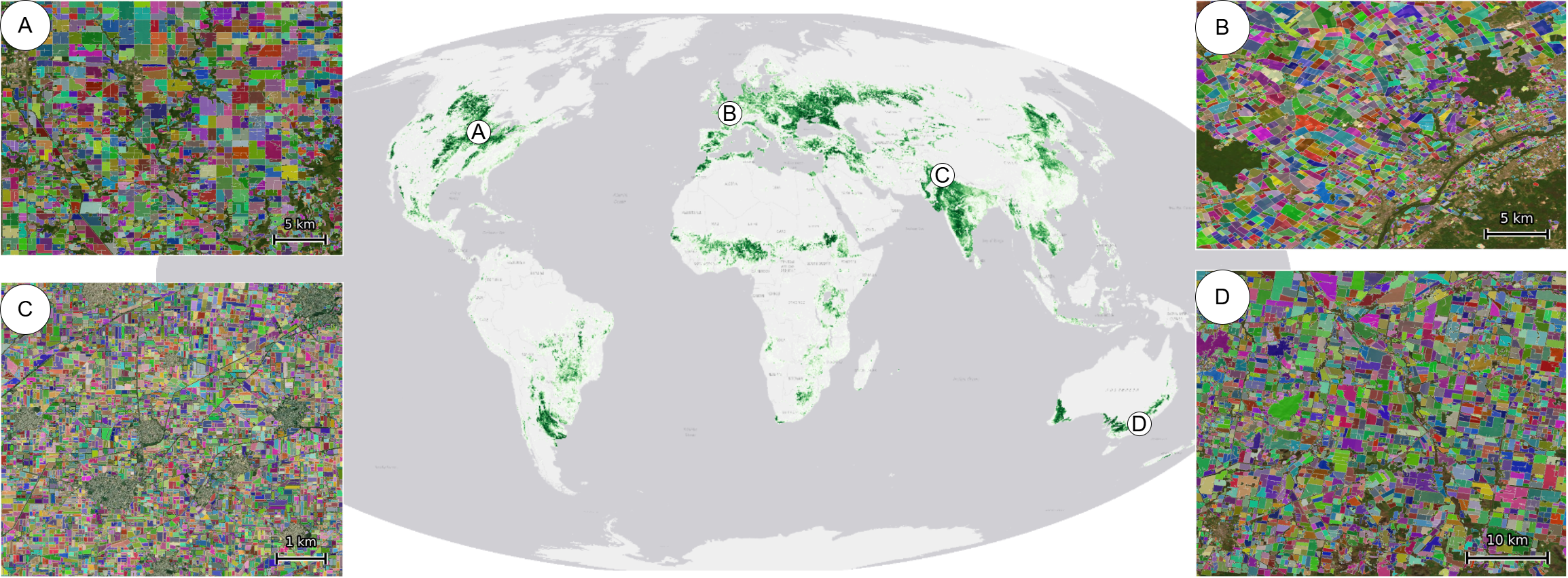}
  \caption{\textbf{Global field boundaries at 10\,m resolution.} The global map shows the total area of our model's predicted fields at 500\,m/px resolution. Insets show predicted field boundaries at 10\,m/px resolution: (a)~Iowa, USA; (b)~Beauce, France; (c)~Punjab, India; (d)~New South Wales, Australia.}
  \label{fig:global_map}
\end{figure*}

Satellite imagery from the Sentinel-2 constellation~\citep{drusch2012sentinel} provides free, global, 10\,m-resolution multispectral imagery with a 5-day revisit, making it an ideal data source for automated, repeatable field boundary extraction.
Recent advances in deep learning for semantic segmentation that have been adapted for remote sensing have shown strong performance on field boundary delineation benchmarks~\citep{waldner2020deep, wang2022unlocking, d2023ai4boundaries}.
Benchmark datasets such as Fields of The World (FTW)~\citep{kerner2025fields}, AI4Boundaries~\citep{d2023ai4boundaries}, AI4SmallFarms~\citep{persello2023ai4smallfarms}, PASTIS~\citep{garnot2021panoptic}, and FBIS-22M~\citep{lavreniuk2025delineate} have accelerated research in this area by providing standardized training and evaluation data across diverse agricultural landscapes.
At the national scale, \citet{estes2022high} produced annual field boundary maps for smallholder-dominated croplands in Ghana, \citet{sadeh2025national} mapped over 5 million fields across Ukraine, \citet{wang2022unlocking} applied transfer learning to smallholder field delineation in India, \citet{rufin2026national} mapped all of Mozambique from SPOT imagery using the DECODE framework, and \citet{lavreniuk2025delineate} demonstrated resolution-agnostic field delineation with zero-shot generalization across geographies.
However, all existing field boundary products are regional in scope, and no study has attempted a truly global, wall-to-wall field boundary map.

Stitching together regional models is also not equivalent to a globally consistent product: per-region models trained on locally available reference data cannot guarantee comparable quality across borders, deliberately stop at country edges, and offer no path forward for the many countries with no public reference data available to train a regional model.
A single model deployed globally enables a single, internally consistent layer that can be compared across borders, and that downstream users do not have to assemble themselves.

Throughout this study, we use the term \textit{field boundary} to mean an observable boundary in satellite imagery separating a contiguous cultivated area from non-cultivated land or from an adjacent cultivated area with distinct spectral or structural properties.
The resulting polygons are \textit{remote-sensing field units} rather than cadastral parcels or ownership-defined management units: depending on field size, landscape structure, and model performance, a polygon may represent a whole field, a sub-field, or a group of adjacent small fields.

We present the first global field boundary map at 10\,m resolution, covering the years 2024 and 2025 (Fig.~\ref{fig:global_map}), with three contributions.
First, we applied the PRUE field boundary segmentation model~\citep{muhawenayo2026prue} to four cloud-free Sentinel-2 mosaics (harvest and planting season snapshots for 2024 and 2025) spanning all land within Sentinel-2 coverage. This resulted in 3.17\,billion remote-sensing field polygons across 241 countries and territories; each polygon is a connected component of the model's predicted field-interior class.
Second, we developed a 500\,m \textit{confidence layer} --- a modeled estimate of whether an area contains true positive predictions --- that indicates how much users should trust the predictions in any given area. The confidence layer achieved an area under the receiver operating characteristic curve (AUC) of 0.82 against ground truth field boundaries from 24 countries using only model-internal features.
Third, we released the full dataset as an open data product: a confidence-thresholded default field boundary map for general users, the full unfiltered dataset for customized use, and the continuous confidence raster for use as a per-cell weight or filterable property. We also provide explicit guidance on the appropriate use of each product.
The complete dataset is released under a CC-BY license in cloud-native formats at \url{https://source.coop/ftw/global-data}, with responsible-use guidance described in this paper.

\subsection*{Modeling approach}

We produced the global field boundary map in three stages: (1) model training, (2) mosaic generation and global inference, and (3) validation and confidence modeling (see Methods for details).

\subsection*{Model training}

We trained the PRUE model~\citep{muhawenayo2026prue}, a U-Net~\citep{ronneberger2015u} with an EfficientNet-B7 encoder~\citep{tan2019efficientnet}, on the CC-BY-licensed subset of the FTW benchmark~\citep{kerner2025fields}, which provides 1.6 million field polygons from 24 countries paired with bi-temporal Sentinel-2 imagery.
The model outputs a three-class semantic segmentation (field interior, field boundary, background) at 10\,m resolution and is designed to reduce tiling artifacts during large-scale deployment~\citep{muhawenayo2026prue}.

\subsection*{Mosaic generation and global inference}

We generated cloud-free harvest and planting season Sentinel-2 global mosaics by selecting scenes with $<$20\% cloud cover covering all land areas between $60^{\circ}$S and $84^{\circ}$N, and computing per-pixel median values across the red, green, blue, and near-infrared bands (B02, B03, B04, B08) at their 10\,m native resolution.
We chose the harvest and planting season temporal windows per $100 \times 100$\,km Sentinel-2 MGRS tile using WorldCereal crop calendar data~\citep{franch2022global} (see Methods).
We processed each global mosaic in overlapping $256 \times 256$ pixel patches, stitched predictions using Gaussian-weighted averaging, and took the argmax of the three-class probability map to produce the final segmentation.
We then extracted individual field instances via connected component analysis on the field interior class.
We parallelized inference across a cluster containing 256 NVIDIA A10G GPUs.

The global inference produced 1.62\,billion individual field polygons for 2024 and 1.55\,billion for 2025, spanning 241 countries and territories.
The unfiltered count includes all connected components with an area of at least 4 pixels ($\approx$ 400 m$^2$ or 0.04 ha).

\begin{figure}[t]
  \centering
  \includegraphics[width=\textwidth]{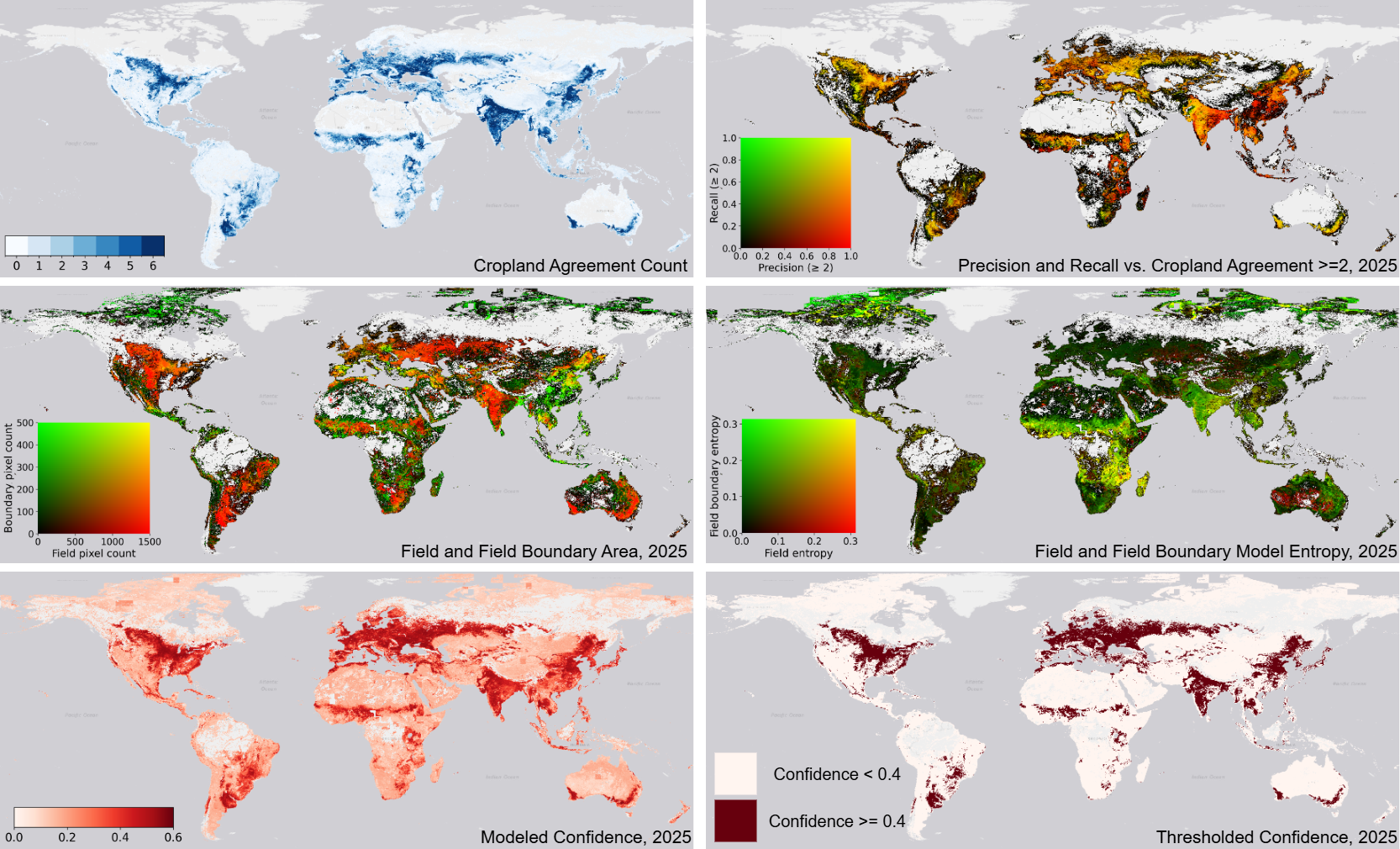}
  \caption{\textbf{500\,m quality indicator rasters and confidence layer.}
  (Top left)~Global cropland agreement map, where each 500\,m pixel is colored by the number of independent global layers that agree the area is cropland.
  (Top right)~Precision and recall of the 2025 field boundaries against the global cropland agreement map at a $\geq$2 threshold.
  (Middle left)~Model-predicted area of field and field boundary classes aggregated to 500\,m/px.
  (Middle right)~Average model entropy over field and field boundary pixels within each 500\,m cell.
  (Bottom left)~Modeled confidence layer.
  (Bottom right)~Thresholded confidence layer used to retain high-confidence predictions in the final filtered product.}
  \label{fig:quality_indicators}
\end{figure}

\subsection*{Validation and confidence layer}

Validation is a fundamental challenge for global data layers derived from machine learning predictions on satellite imagery.
Random sampling and labeling of model outputs can provide an estimate of precision (the fraction of predicted fields that are true fields), but not recall (the fraction of true fields that are predicted as fields): if we already knew where all the fields were, there would be no need to produce a global map in the first place.
Annotating field boundaries at 10\,m resolution is also not a task that can be reliably crowdsourced, as an annotator must recognize regional cropping calendars, distinguish annual crop fields from pasture, orchards, and fallow, and resolve pixel-scale boundaries between adjacent management units --- skills that require training in agriculture and remote-sensing interpretation rather than general visual judgment~\citep{see2013comparing}. Prior large-scale crowdsourcing of agricultural land-cover and field-size labels has consequently relied on trained contributors, expert validation, or both to produce usable results~\citep{fritz2015mapping, lesiv2019estimating}.
Further, global models inevitably have regions where the input imagery is out-of-distribution relative to the training data, where cloud contamination degrades inputs, or where the target agricultural landscapes differ from anything in the training set~\citep{rolf2026contrasting}.

Therefore, we combined three complementary evaluations: (1) pixel-level recall against the full FTW ground-truth polygon set across 24 countries, (2) full-country precision and recall against national LPIS/INVEKOS parcel databases in Austria, Latvia, and Finland, and (3) a 500\,m \textit{confidence layer} trained to predict cell-level reliability everywhere on the globe, including in the majority of countries where no reference data exists. The first two quantify model accuracy where ground truth is available; the third extrapolates reliability to the regions where it is not.

The confidence layer is a 500\,m raster indicating how much users should trust the field predictions in a given area (Fig.~\ref{fig:quality_indicators}).
A high confidence value means the model produces field boundary predictions in a raster cell that actually contains fields; a low value means either that there are no fields, or that the model's predictions there resemble spurious patterns it produces over forests, deserts, or wetlands (i.e. are false positives).
This layer has two purposes: (1) it provides a default filtered view of the dataset that removes obvious false positives, and (2) it flags regions where the model may have blind spots, allowing users to assess prediction reliability in their area of interest.
We released the full unfiltered dataset for users who wish to apply their own quality criteria.
After filtering at the default conf\,$\geq$\,0.4 threshold, 864\,million fields were retained in 2024 and 844\,million (54.3\%) in 2025.

\paragraph{Comparison with labeled data.}
We computed pixel-level recall over the full set of ground-truth field boundary polygons from which the FTW training samples were derived, covering 24 countries.
For each country, we rasterized the field polygons to 10\,m/px binary masks where pixels are labeled as field or background.
We then rasterized the global 2025 predictions to the same grid and computed recall at the pixel level.

\begin{figure}[!ht]
  \centering
  \includegraphics[width=0.49\linewidth]{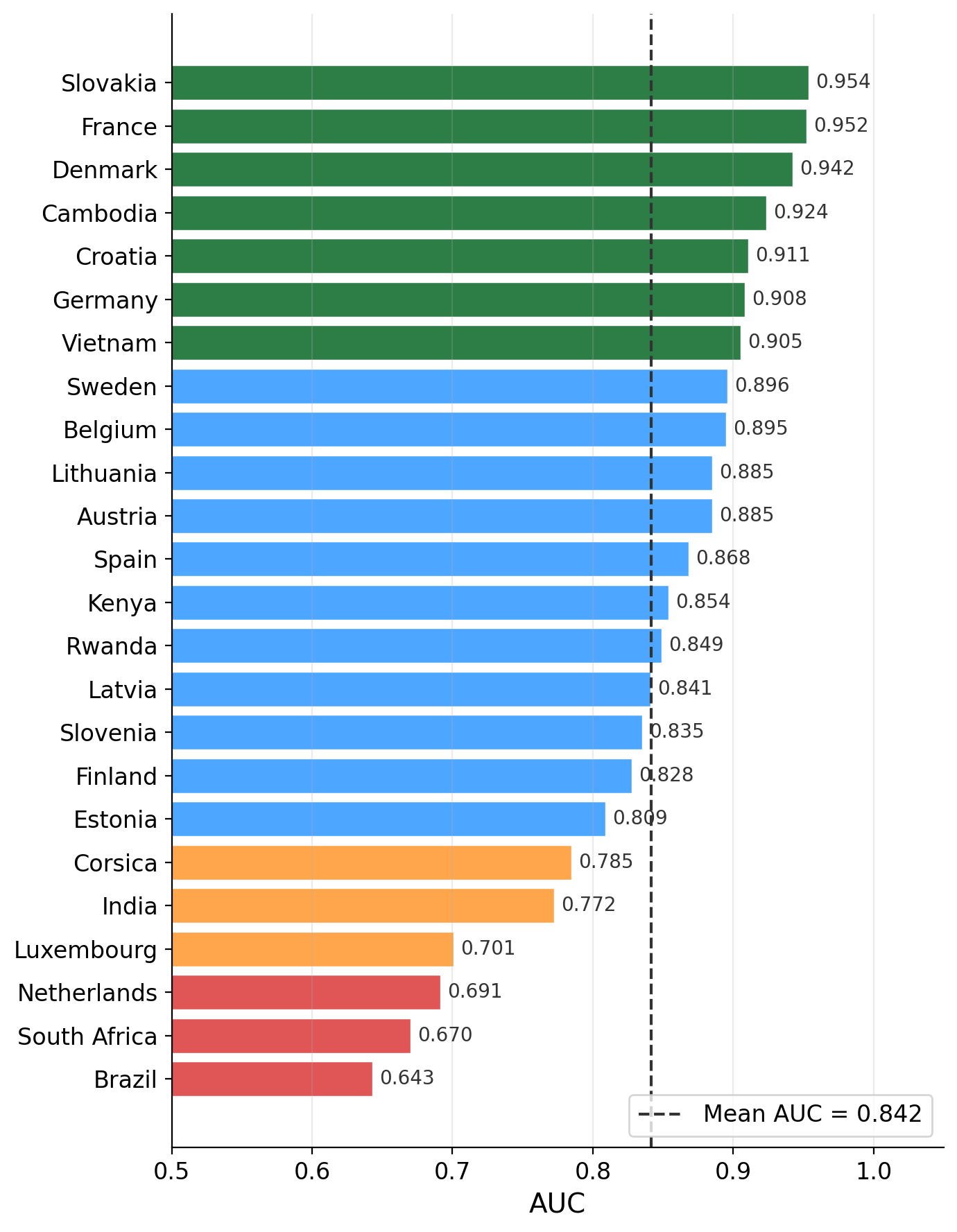}
  \includegraphics[width=0.49\linewidth]{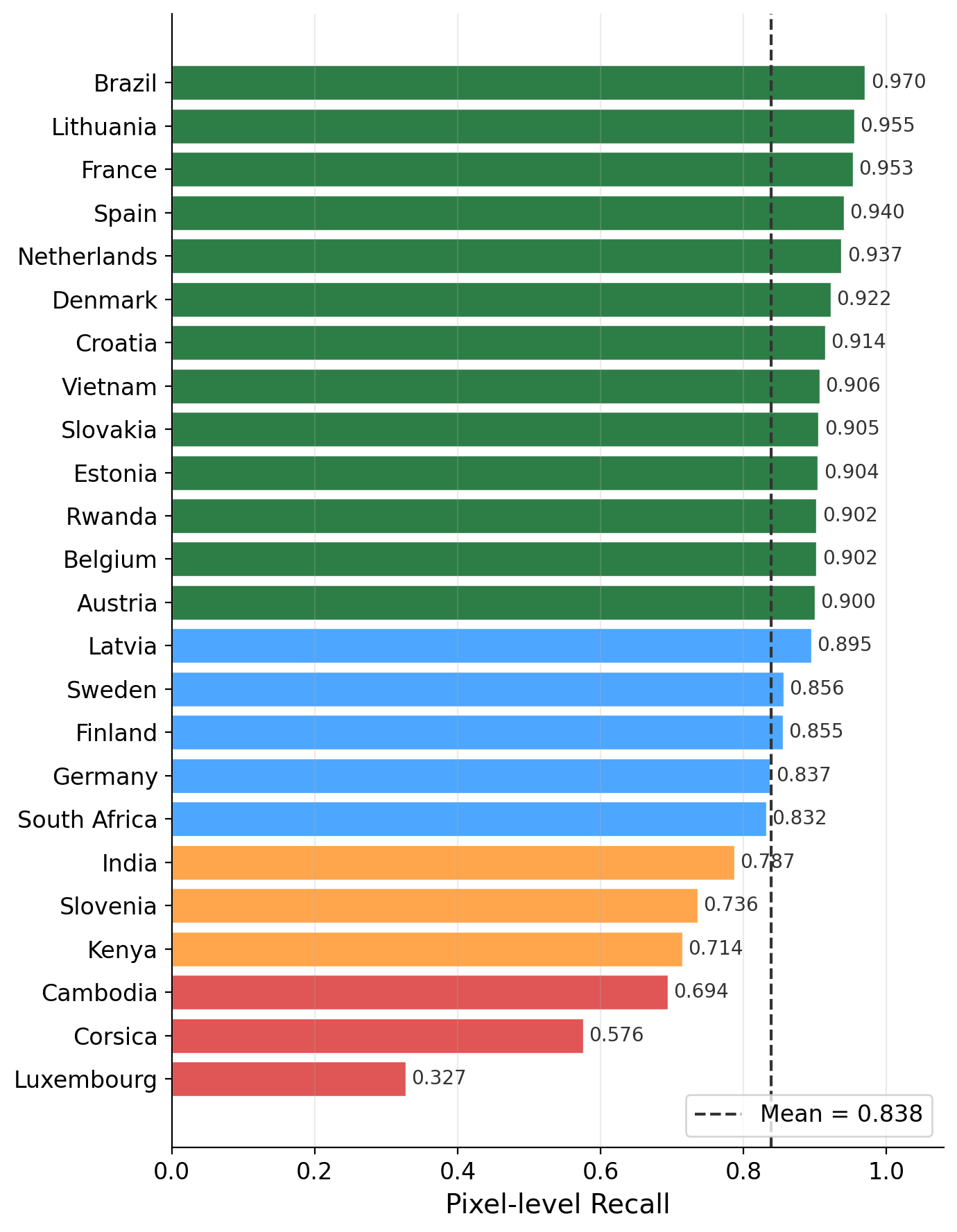}
  \caption{\textbf{Per-country confidence-model transferability and pixel-level recall.}
  (Left)~Leave-one-country-out AUC for the Random Forest confidence model; the dashed line shows the unweighted mean AUC (0.842).
  (Right)~Per-country pixel-level recall of PRUE 2025 global predictions against ground-truth field boundaries for 24 countries; the dashed line shows the mean recall (0.852).
  Portugal is excluded from both panels due to insufficient reference data (Methods).}
  \label{fig:per_country_results}
\end{figure}

The right panel of Fig.~\ref{fig:per_country_results} reports per-country recall results. Recall exceeded 0.90 for 14 of 24~evaluated countries, with a mean of 0.852 across countries. The highest recall was observed in Brazil (0.970), Lithuania (0.955), and France (0.953). Lower recall in Luxembourg (0.327), Corsica (0.576), and Cambodia (0.694) likely reflects regional landscape heterogeneity or ground-truth misalignment rather than systematic model failure. Portugal was excluded entirely: its ground truth data comes from two islands in the Azores with only 5\,040 polygons, and the model predicted no fields there (recall\,=\,0). Four countries (Brazil, India, Kenya, Rwanda) used presence-only ground truth, where non-field labels are inferred from the absence of field polygons rather than explicit annotation; this does not affect recall computation (which depends only on field pixels) but does impact the confidence model evaluation.

\paragraph{Confidence model.}
The confidence model predicts whether each 500\,m cell in the global map contains true positive field predictions (see Methods for details). To create input features for this model, we computed 500\,m-resolution summary statistics from the 10\,m-resolution model outputs: model entropy per class, predicted area per class, and agreement with eight independent global cropland products. We used these features to train a Random Forest classifier on 500\,m cells drawn from the 24 FTW-labeled countries, labeling each cell as positive if it overlaps at least one FTW ground-truth field polygon.
Formally, the per-cell confidence score is the classifier's estimated posterior probability $P(\text{true-positive field in cell} \mid \text{model-derived features})$: values near 1 indicate the model's predictions in that cell statistically resemble predictions observed over ground-truth labeled fields, and values near 0 indicate they resemble predictions over areas the reference data and the external cropland consensus both agree are not cropland.
A key training challenge is obtaining reliable negative examples: ground-truth coverage is often incomplete so a cell without a labeled field is not necessarily a non-field cell.
We addressed this by retaining as negatives only cells where both the ground truth and an independent consensus of eight global cropland layers agree the area is not cropland (consensus count $\leq 2$).
With the resulting training dataset, the confidence model achieved AUC = 0.82 in 5-fold cross-validation using only model-internal features (Table~\ref{tab:confidence_model}). Leave-one-country-out evaluation confirmed geographic transferability with a mean AUC of 0.84 and per-country values ranging from 0.64 (Brazil) to 0.95 (Slovakia) (Fig.~\ref{fig:per_country_results}).
Including external cropland consensus features in addition to the model-internal features raised the 5-fold AUC to 0.96. However, this also introduced some confirmation bias, since the cropland consensus used as a training feature was also used to construct the negative labels. We therefore defaulted to the model-only configuration in the released confidence raster, since it reflects PRUE's own quality rather than the cropland consensus.
The score is a cell-level reliability estimate, not a measure of the geometric accuracy of any individual polygon boundary within the cell; downstream users should not infer polygon-level geometric fidelity from cell-level confidence scores. Object-level evaluation against independent reference data remains a priority for future work~\citep{radoux2017good, ye2018review}.

\begin{table}[!ht]
\centering
\caption{\textbf{Confidence model performance} (5-fold cross-validation after conservative filtering, crop consensus $\leq 2$).
``Model-only'' features use only entropy and prediction density (no external cropland data); ``Model+P/R'' additionally includes precision/recall against the cropland consensus.
See Methods for unfiltered baselines, additional filter thresholds, and feature configurations.}
\label{tab:confidence_model}
\small
\begin{tabular}{llcccc}
\toprule
\textbf{Features} & \textbf{Model} & \textbf{AUC} & \textbf{F1} & \textbf{Precision} & \textbf{Recall} \\
\midrule
Model-only   & Logistic Regression & 0.75$\pm$0.00 & 0.71$\pm$0.01 & 0.70$\pm$0.01 & 0.71$\pm$0.01 \\
Model-only   & Random Forest       & 0.82$\pm$0.00 & 0.76$\pm$0.00 & 0.74$\pm$0.00 & 0.78$\pm$0.01 \\
\addlinespace
Model+P/R   & Logistic Regression & 0.94$\pm$0.00 & 0.87$\pm$0.00 & 0.86$\pm$0.01 & 0.89$\pm$0.00 \\
Model+P/R   & Random Forest       & 0.96$\pm$0.00 & 0.91$\pm$0.00 & 0.98$\pm$0.00 & 0.86$\pm$0.00 \\
\bottomrule
\end{tabular}
\end{table}

\begin{figure*}[!ht]
  \centering
  \includegraphics[width=\textwidth]{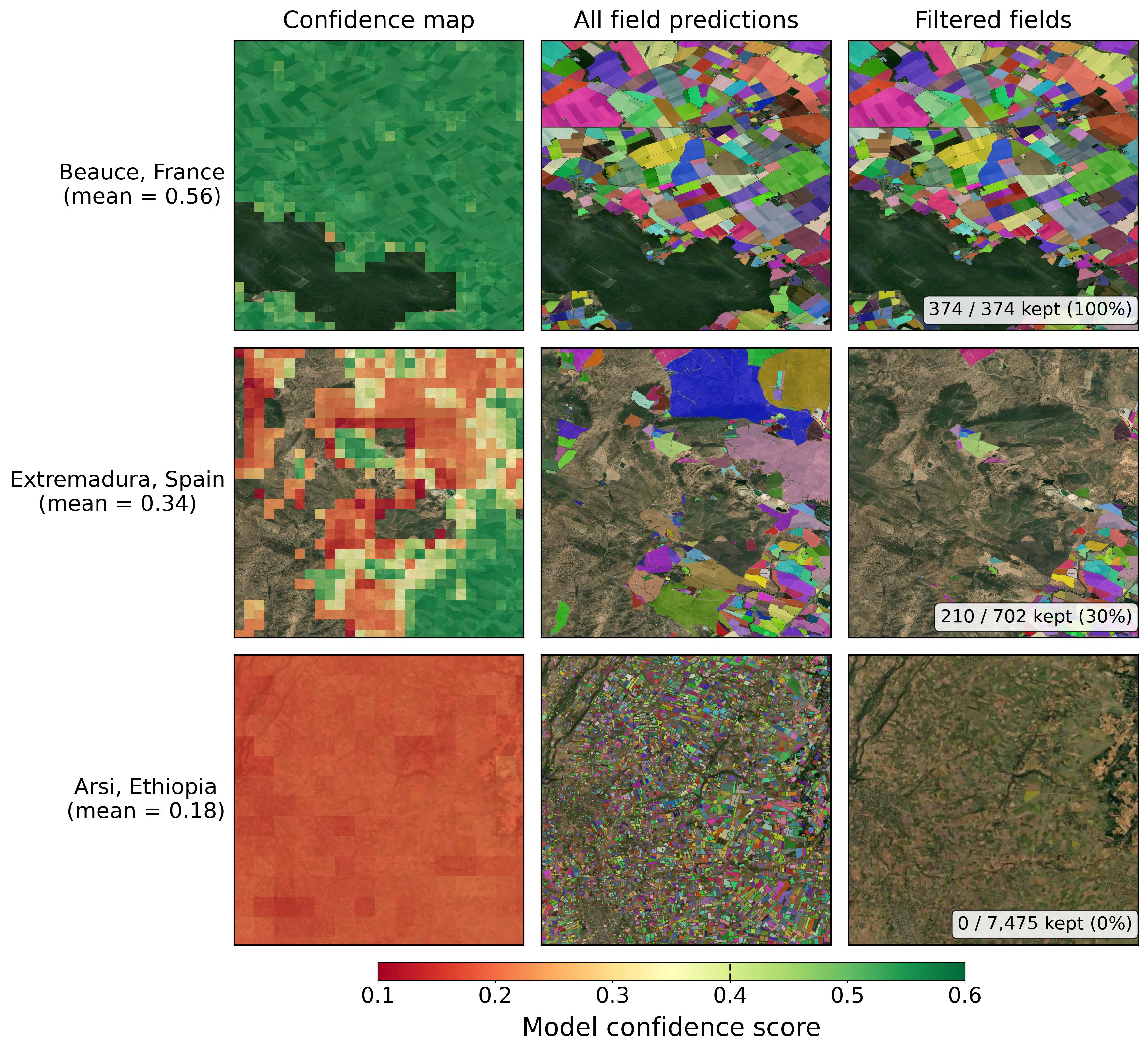}
  \caption{\textbf{Confidence layer behavior at three sites with high, medium, and low modeled confidence.}
  Rows: (Top) Beauce, France (a country well represented in FTW training data); (Middle) Extremadura, Spain, with intermediate modeled confidence; (Bottom) Arsi, Ethiopia, a region outside FTW's training coverage.
  Columns: (Left)~the 500\,m confidence raster; (Center)~all PRUE field polygons (unfiltered); (Right)~polygons retained at the recommended conf\,$\geq$\,0.4 threshold.
  At this threshold, 100\% of polygons are retained at the Beauce site, $\sim$30\% at the Extremadura site, and 0\% at the Arsi site. We observe that the unfiltered Arsi predictions visually correspond to genuine smallholder fields and note that the confidence layer is conservative in regions whose visual signature differs from those seen during training.}
  \label{fig:confidence_examples}
\end{figure*}

\paragraph{Confidence layer in practice.}
Figure~\ref{fig:confidence_examples} illustrates how the confidence layer translates into different operational outcomes at three sites representing high, medium, and low modeled confidence.
In the Beauce region of France (a country well represented in FTW training data), confidence values were uniformly high and the default filter retained 100\% of predicted polygons.
In Extremadura, Spain, confidence values were mixed and the default filter retained roughly 30\% of polygons.
In the Arsi highlands of Ethiopia, a region outside FTW's training coverage, the confidence layer assigned uniformly low values and the default filter removed all polygons, even though the unfiltered predictions visually correspond to genuine smallholder fields.
This is the conservative-by-design behavior of the confidence layer: it does not certify predictions in regions whose visual signature differs from those seen during training, which means real fields in underrepresented smallholder systems may be filtered out.
Users working in such regions should examine the unfiltered product and the continuous confidence raster directly rather than relying on the default threshold.

\begin{table}[!ht]
\centering
\caption{\textbf{Full-country pixel-level validation of PRUE 2024 predictions against national LPIS/INVEKOS seasonal-crop masks.}
All metrics are computed at 10\,m resolution within the ADM0 national boundary after restricting the ground truth to seasonal (annual) crops, which matches FTW's training scope~\citep{kerner2025fields}.
Permanent grassland, pasture, orchards, vineyards, fallow, and forestry parcels are excluded from the ground truth because the model is not trained to detect them; any PRUE pixel falling on such parcels therefore contributes to a false positive.
``conf\,$\geq$\,0.4'' applies the recommended confidence threshold from the 500\,m confidence layer.}
\label{tab:full_country_metrics}
\small
\begin{tabular}{llcccc}
\toprule
\textbf{Country} & \textbf{Version} & \textbf{Precision} & \textbf{Recall} & \textbf{F1} & \textbf{IoU} \\
\midrule
Austria (INVEKOS 2024)  & Unfiltered        & 0.900 & 0.884 & 0.892 & 0.805 \\
                        & conf\,$\geq$\,0.4 & 0.904 & 0.865 & 0.884 & 0.792 \\
\addlinespace
Latvia (LPIS 2024)      & Unfiltered        & 0.870 & 0.898 & 0.884 & 0.792 \\
                        & conf\,$\geq$\,0.4 & 0.883 & 0.858 & 0.870 & 0.771 \\
\addlinespace
Finland (LPIS 2024)     & Unfiltered        & 0.650 & 0.860 & 0.740 & 0.588 \\
                        & conf\,$\geq$\,0.4 & 0.703 & 0.792 & 0.745 & 0.594 \\
\bottomrule
\end{tabular}
\end{table}

\paragraph{Full-country pixel-level validation.}
The per-country recall analysis above is restricted to DBSCAN-derived coverage hulls around FTW ground-truth samples and does not report precision.
To complement it with a full-country precision and recall assessment, we evaluated the PRUE 2024 predictions against complete national Land Parcel Identification System (LPIS) and INVEKOS databases for Austria, Finland, and Latvia (Table~\ref{tab:full_country_metrics}).
Because FTW was trained exclusively on annual crops and explicitly excludes permanent grassland, pasture, orchards, vineyards, fallow, and forestry~\citep{kerner2025fields}, we filtered the national reference data to the same seasonal-crop scope before rasterizing and comparing against the model's field and field-boundary classes at 10\,m resolution (see Methods).
The unfiltered F1 reached 0.89 in Austria (precision 0.90, recall 0.88) and 0.88 in Latvia (precision 0.87, recall 0.90).
Finland was substantially harder (F1\,=\,0.74; precision 0.65, recall 0.86): performance degraded with latitude, from F1\,=\,0.83 in the southern agricultural belt (60$^{\circ}$\,N) to F1\,$<$\,0.13 in Lapland (66$^{\circ}$\,N), where the model over-predicted fields on boreal forest clearings, bogs, and wet meadows.
The default filter modestly increased precision in all three countries at a small cost in recall; in Finland, for example, precision rose from 0.65 to 0.70 while recall dropped from 0.86 to 0.79, leaving F1 essentially unchanged (0.74 to 0.75).
The Finland gradient was consistent with the confidence layer assigning lower scores at high latitudes and illustrated that model failure modes outside the FTW training distribution are captured by the confidence layer even where mean F1 is substantially reduced.

\begin{table}[!ht]
\centering
\caption{\textbf{Distributional comparison of PRUE 2025 field polygons against an independent ML-derived reference dataset over Zambia.}
Neither product is ground truth; the comparison characterizes systematic differences rather than accuracy.
PRUE statistics are reported both unfiltered and at the recommended confidence threshold (conf\,$\geq$\,0.4).}
\label{tab:zambia_distribution}
\small
\resizebox{\linewidth}{!}{%
\begin{tabular}{lrrr}
\toprule
\textbf{Metric} & \textbf{Reference 2024} & \textbf{PRUE 2025 (all)} & \textbf{PRUE 2025 (conf\,$\geq$\,0.4)} \\
\midrule
Field polygons ($\times10^{6}$) & 7.7 & 39.4 & 6.9 \\
Total mapped area (Mha)         & 9.2 & --- & 1.9 \\
Median area (ha)                & 0.31 & 0.06 & 0.07 \\
Median perimeter (m)            & 247  & 118 & 119 \\
Median compactness (Polsby--Popper) & 0.67 & 0.59 & 0.58 \\
Median fractal dimension        & 1.38 & 1.05 & 1.05 \\
\bottomrule
\end{tabular}%
}
\end{table}

\paragraph{Distributional comparison in Zambia.}
Public ground-truth field boundaries are not available for most smallholder regions, so direct validation there is not possible.
We instead compared the PRUE 2025 Zambia polygons to an independent 2024 ML-derived field boundary dataset for Zambia, produced by a PRUE-variant model~\citep{muhawenayo2026prue} (with different loss and normalization) trained on the Mapping Africa Planet-based reference data and applied to Planet NICFI tiles~\citep{estes2024region, khallaghi2025generalization} (Table~\ref{tab:zambia_distribution}); neither product is ground truth, so this characterizes systematic differences, not accuracy.
At the default filter, PRUE yielded 6.9\,million polygons versus 7.7\,million in the reference, but mapped only 1.9\,Mha of total field area versus 9.2\,Mha.
Median PRUE field area was 0.07\,ha versus 0.31\,ha, and median boundary fractal dimension was 1.05 versus 1.38 (values close to 1.0 indicate boundaries that trace the 10\,m pixel grid rather than the natural, curved edges of real fields).
PRUE therefore fragmented individual smallholder fields into clusters of pixel-scale polygons, producing comparable polygon counts but much less total mapped area and lower polygon-level geometric fidelity.
We flagged over-fragmentation as a known failure mode in smallholder systems and a priority for future work on post-processing or higher-resolution retraining.

\subsection*{Discussion}

The global field boundary map and accompanying confidence layer address a gap that has persisted despite growing demand for field-level agricultural data: no prior dataset provides wall-to-wall coverage or spatially explicit quality information.

\paragraph{Why a single global, open product matters.}
Previous field boundary maps cover individual countries~\citep{estes2022high, sadeh2025national,rufin2026national} or continents~\citep{d2023ai4boundaries}, and the highest-quality boundary data remain locked inside national LPIS systems and proprietary corporate datasets. This distribution has largely restricted field-level agricultural science at scale to institutions with access to those sources.
Our product is the first to cover all global cropland areas with a single internally consistent model, released under a CC-BY license in the Field Boundaries for Agriculture (fiboa)\footnote{\url{https://fiboa.org}} standard.
Open licensing lowers the practical barrier to field-scale analysis for institutions that lack both LPIS access and the resources to commission or license equivalent data, including national statistics agencies in LPIS-absent jurisdictions, academic groups, and NGOs.

\paragraph{From pixel-level to field-level agriculture.}
Beyond the dataset itself, this work provides a globally consistent unit of analysis that matches how agriculture is organized.
Most existing global agricultural products---cropland masks, vegetation indices, NDVI-derived yield proxies---operate at the pixel level; operational programs such as the G20 GEOGLAM Crop Monitor for AMIS synthesize these into consensus country- and regional-scale assessments underpinning international food-market transparency~\citep{becker2019geoglam}, but neither a pixel nor an administrative region is the unit at which individual management decisions occur.
A global field-level layer changes what is computable: aggregate farm structure (median field size by continent, distribution tails), boundary-aware crop type mapping that pools pixels within fields, field-level MRV for conservation programs, change detection for consolidation and fragmentation, and stratified survey design in countries that previously had no spatial frame.
These analyses do not require a perfect product, but they do require an open, globally available one with spatially explicit quality information.

\paragraph{Confidence layer as a tiered data product.}
Releasing the confidence layer alongside the raw predictions enables a tiered access model: general users can work with the default filtered product (threshold $\geq 0.4$ retains 844\,M fields), applications requiring high precision, such as area estimation for policy reporting, can apply a stricter threshold ($\geq 0.5$ removes roughly 25\% of active cells), and downstream analyses can use the continuous confidence raster as a per-cell weight rather than a binary filter.
This tiered design lets users choose the precision/recall tradeoff appropriate for their application, rather than committing everyone to one.

\paragraph{Limitations.}
First, the model inherits FTW's training scope: annual field crops only, excluding pasture, perennial crops, orchards, vineyards, and other tree crops~\citep{kerner2025fields}, so these classes are missed systematically.
Geographic coverage is heavily European (17 of 24 countries), with four countries (Brazil, India, Kenya, Rwanda) providing only sparse presence-only labels; North America, China, Russia and Central Asia, Australia, the Middle East, and most of North Africa are entirely absent from training, as are paddy rice outside Cambodia and Vietnam, semi-arid and arid irrigated agriculture, terraced highland smallholder systems, and boreal cropping. The Finnish boreal over-prediction and Zambian smallholder over-fragmentation described above are concrete examples of these gaps.
Second, the negative-label filter relies on a cropland consensus that may share systematic biases across its eight constituent layers; a probabilistic framework for partial reference data~\citep{olofsson2014good} would provide more rigorous accuracy estimates.
Third, full-country precision is reported only for Austria, Latvia, and Finland; a globally stratified random-sampling campaign to estimate precision across agro-climatic zones remains a priority.
Fourth, the 500\,m confidence indicators cannot capture polygon-level geometric accuracy~\citep{stehman2011pixels}, and the Zambia comparison shows that the model can produce clusters of pixel-scale polygons where a single smallholder field exists; object-level evaluation~\citep{radoux2017good, ye2018review} against independent reference data is needed.

\paragraph{Responsible use.}
This dataset is intended for research, monitoring, and analytical applications at scale.
\textbf{It is not a cadastral or land-tenure product.}
Polygon boundaries follow Sentinel-2 pixel edges rather than legal parcel boundaries, and a single legal parcel may correspond to many polygons, none, or a cluster that crosses into neighboring parcels.

\paragraph{Outlook.}
Future directions include the stratified manual precision audit described above, direct object-level accuracy assessment~\citep{olofsson2014good, radoux2017good, ye2018review} using stratified random sampling, temporal analysis of field boundary change between the 2024 and 2025 maps, post-processing to address the smallholder over-fragmentation surfaced by the Zambia comparison, and integration with downstream applications such as crop type mapping and yield estimation.

\subsection*{Data availability}

The global field boundary product (2024 and 2025), 500\,m quality indicator rasters, confidence model artifacts, and confidence-filtered field density layers are available at \url{https://source.coop/ftw/global-data} under a CC-BY license.
See Methods for a full description of all released data layers.
The Fields of The World training data is available at \url{https://source.coop/kerner-lab/fields-of-the-world}.

\subsection*{Code availability}

Model training code and the FTW baseline implementations are available at \url{https://github.com/fieldsoftheworld/ftw-baselines}.

\subsection*{Acknowledgements}
This work was supported by academic research grant funding and technical support resources provided by Taylor Geospatial.
ZF was supported by funding from NASA's Land Cover Land-Use Change program, award \#80NSSC23K0528.
G. Essuman provided the independent Zambia 2024 field boundary dataset and associated shape metrics.

\bibliographystyle{plainnat}
\bibliography{citations}

\clearpage
\setcounter{figure}{0}
\setcounter{table}{0}
\renewcommand{\thefigure}{M\arabic{figure}}
\renewcommand{\thetable}{M\arabic{table}}

\section*{Methods}

\subsection*{Training data}

We trained the model on the CC-BY-licensed subset of the Fields of The World (FTW) dataset~\citep{kerner2025fields}.
FTW provides over 70,000 training samples spanning 24 countries, each consisting of a $256 \times 256$ pixel patch of bi-temporal Sentinel-2 RGBN imagery paired with instance and semantic segmentation masks.
We used FTW's predefined train/validation/test splits, designed to minimize spatial autocorrelation.
Following \citet{kerner2025fields}, we masked pixels with unknown labels during training for presence-only examples.

\subsection*{Model architecture and training}

We used the PRUE model~\citep{muhawenayo2026prue}, which combines a U-Net~\citep{ronneberger2015u} decoder with an EfficientNet-B7~\citep{tan2019efficientnet} encoder (67.1\,M parameters).
The encoder processes the 8-channel bi-temporal input (4 RGBN bands $\times$ 2 time steps) after adapting the input convolution layer.
We trained with the Adam optimizer, log-cosh Dice loss, and class weights $[0.05, 0.20, 0.75]$ for background, field interior, and boundary, respectively (boundary weight $\omega = 0.75$).
Data augmentations included channel shuffling for input-order invariance, random brightness jittering, and random resize for scale robustness.
See \citet{muhawenayo2026prue} for full training details including learning rate selection and hyperparameter sweeps.

\subsection*{Mosaic generation}

We generated cloud-free Sentinel-2 Level-2A composites for two seasons per year (planting and harvest).
To assign season windows per tile, we used the WorldCereal global crop calendar~\citep{franch2022global}, which provides gridded start-of-season (SOS) and end-of-season (EOS) rasters at 0.5$^{\circ}$ resolution for wheat and maize.
For each Sentinel-2 MGRS tile, we sampled the SOS and EOS rasters to define a planting window bracketing SOS and a harvest window bracketing EOS.
For each tile and season, we selected all available scenes with cloud cover $< 20\%$, computed per-pixel median composites, and applied a final cloud mask using the Sentinel-2 SCL band.
We generated composites for the RGBN bands (B02, B03, B04, B08) at 10\,m native resolution.

\subsection*{Inference pipeline}

We ran global inference tile-by-tile across the ESA WorldCover grid.
For each tile, we divided the bi-temporal mosaic into $256 \times 256$ pixel patches with 25\% overlap (64-pixel stride on each edge).
We normalized each patch to surface reflectance units (division by 10,000 with BOA offset correction) and passed it through the model to produce a $256 \times 256 \times 3$ probability map.
We combined overlapping predictions using Gaussian-weighted averaging: a 2D Gaussian kernel ($\sigma = 0.25 \times 256$) centered on each patch assigns higher weight to central predictions.
We took the argmax of the final stitched probability map to produce a three-class raster (background, field interior, boundary).
Each polygon in our dataset is extracted as a connected component of the predicted field-interior class after segmentation.
Thus, the map delineates visually separable cultivated units as resolved by Sentinel-2 imagery and the PRUE model, rather than legal parcels, ownership units, or farm-management units.
Internal crop, tillage, irrigation, soil, or phenological differences within a large managed field may produce multiple mapped polygons, whereas adjacent small fields may be merged when their separating boundaries are not resolved at 10\,m resolution.
The resulting polygons are serialized as fiboa-compliant GeoParquet files.

\subsection*{Quality indicator computation}

We computed the 500\,m raster layers from the full-resolution (10\,m) outputs on a common global grid: 86,400 $\times$ 34,560 pixels in EPSG:4326, covering $-180^{\circ}$ to $180^{\circ}$ longitude and $-60^{\circ}$ to $84^{\circ}$ latitude ($\sim$0.00417$^{\circ}$/pixel).
Each 500\,m cell corresponds to approximately $50 \times 50$ 10\,m pixels.

\begin{description}[style=unboxed]
  \item[Entropy.] For each 500\,m cell, we computed the Shannon entropy $H = -\sum_c p_c \log p_c$ per 10\,m pixel separately for the argmaxed field and field boundary pixels, then averaged across the cell.
  \item[Prediction density.] We counted the number of 10\,m pixels where the argmax of the three-class probabilities yielded ``field'' or ``boundary''.
  \item[Precision and recall.] For each 500\,m cell we compared the set of 10\,m pixels predicted as field by the model ($F$) against the set of 10\,m pixels identified as cropland by the consensus layer ($C_k$), where $C_k$ denotes pixels for which at least $k$ of the eight independent cropland datasets agree.
    We computed precision and recall at two agreement thresholds ($k \in \{2, 3\}$):
    \[
      \text{Precision}_k = \frac{|F \cap C_k|}{|F|}, \qquad
      \text{Recall}_k = \frac{|F \cap C_k|}{|C_k|}
    \]
    where all set operations are restricted to pixels within the 500\,m cell.
    High precision indicates that predicted fields coincide with independently mapped cropland; high recall indicates that the model captures most of the consensus cropland area.
  \item[Crop consensus count.] We aggregated eight independent global cropland datasets into a per-pixel agreement score. Specifically, we reprojected and resampled (with nearest neighbor interpolation) each dataset to the ESA WorldCover 10\,m grid and binarized according to dataset-specific cropland classes. We then summed per-pixel across all eight layers and averaged within each 500\,m cell to produce a continuous consensus value. Outside Africa, the practical maximum is~7 since Digital Earth Africa provides coverage for the African continent only. Table~\ref{tab:cropland-layers} lists each contributing dataset and the respective cropland classes we included.
\end{description}

\begin{table}[!ht]
\centering
\caption{\textbf{Global cropland datasets} merged into the consensus agreement layer. Each dataset is binarized to cropland/non-cropland at the listed class values and reprojected to the ESA WorldCover 10\,m grid.}
\label{tab:cropland-layers}
\scriptsize
\begin{tabular}{llll}
\toprule
\textbf{Dataset} & \textbf{Resolution} & \textbf{Cropland classes} & \textbf{Reference} \\
\midrule
ASAP Crop Mask v04
  & 500\,m & Value $> 0$ (fractional cover)
  & \cite{rembold2019asap} \\
ESA GlobCover 2009
  & 300\,m & 11, 14, 20, 30
  & \cite{arino2012global} \\
ESA CCI Land Cover 2020
  & 300\,m & 10, 11, 12, 20, 30, 40
  & \cite{c3s_landcover} \\
Copernicus Global LC 100\,m v3
  & 100\,m & Class 40
  & \cite{buchhorn2020copernicus} \\
GLAD Cropland 2019
  & 30\,m & Class 1
  & \cite{potapov2022global} \\
Esri 10\,m LULC 2021
  & 10\,m & Class 5
  & \cite{karra2021global} \\
Digital Earth Africa 2019
  & 10\,m & Class 1 (Africa only)
  & \cite{burton2022co} \\
ESA WorldCereal 2021
  & 10\,m & Class 100
  & \cite{van2023worldcereal} \\
\bottomrule
\end{tabular}
\end{table}

\paragraph{Confidence model feature sets.}
From these indicators we derived three additional per-cell ratios: field-to-boundary pixel ratio, field density (fraction of cell classified as field), and entropy ratio (field entropy divided by boundary entropy).
To disentangle model-internal signals from external cropland references and to test for circularity with the consensus-based negative filtering, we grouped features into four configurations (Table~\ref{tab:feature_sets}).

\begin{table}[!ht]
\centering
\caption{\textbf{Confidence model feature sets.} ``Model-only'' uses only PRUE outputs and is free of circularity with the crop-count filter. Adding external cropland features improves AUC but introduces partial circularity.}
\label{tab:feature_sets}
\small
\begin{tabular}{lcp{7.5cm}}
\toprule
\textbf{Feature set} & \textbf{$n$} & \textbf{Constituents} \\
\midrule
Model-only & 7 & Entropy (field, boundary), pixel count (field, boundary), field-to-boundary ratio, field density, entropy ratio \\
Model+consensus & 8 & Model-only + crop consensus count \\
Model+P/R & 11 & Model-only + precision and recall vs.\ cropland consensus at $\geq$\,2 and $\geq$\,3 agreement thresholds \\
All & 12 & Model+P/R + crop consensus count \\
\bottomrule
\end{tabular}
\end{table}

\subsection*{Data products}

\begin{figure}[t]
  \centering
  \includegraphics[width=0.49\linewidth]{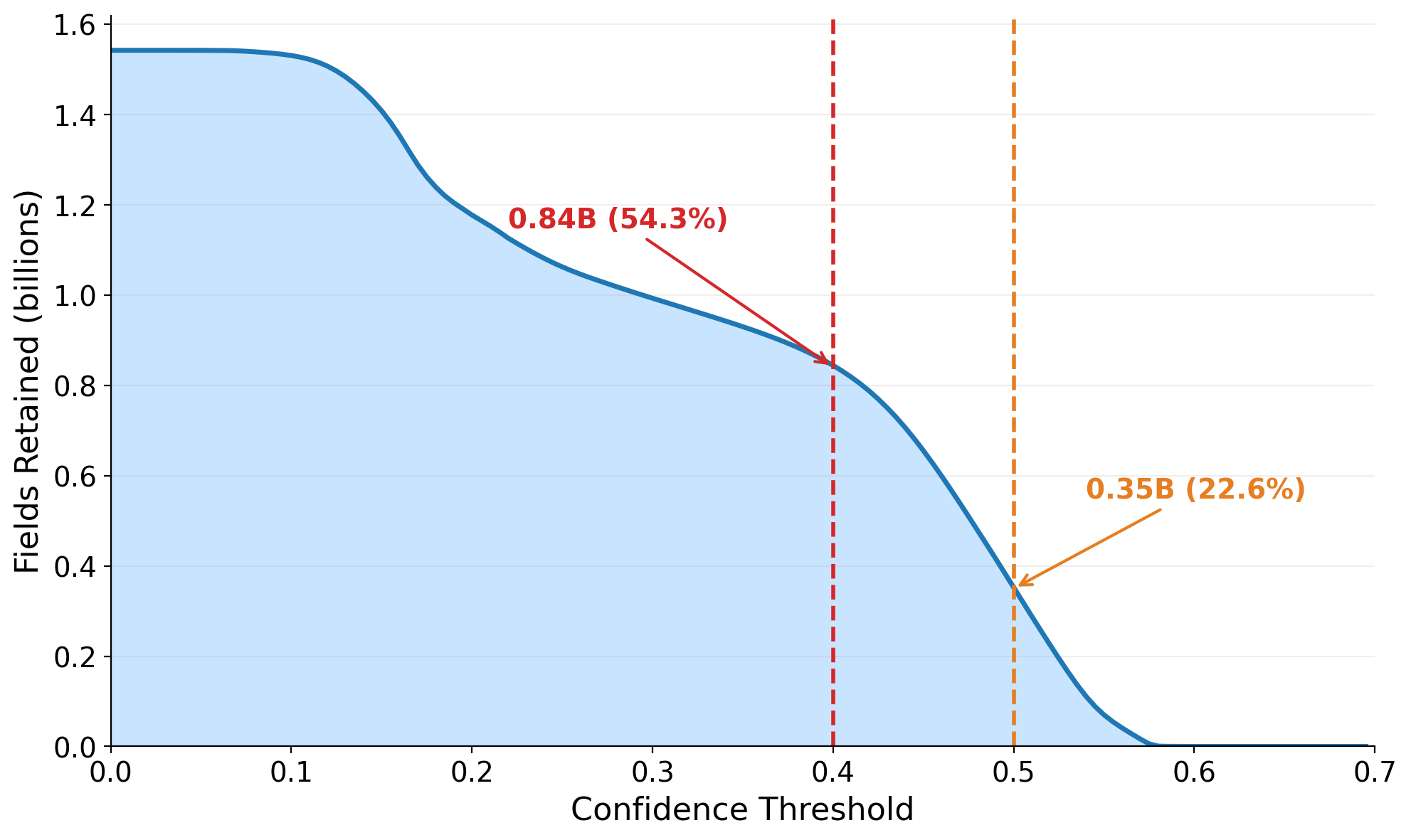}
  \includegraphics[width=0.49\linewidth]{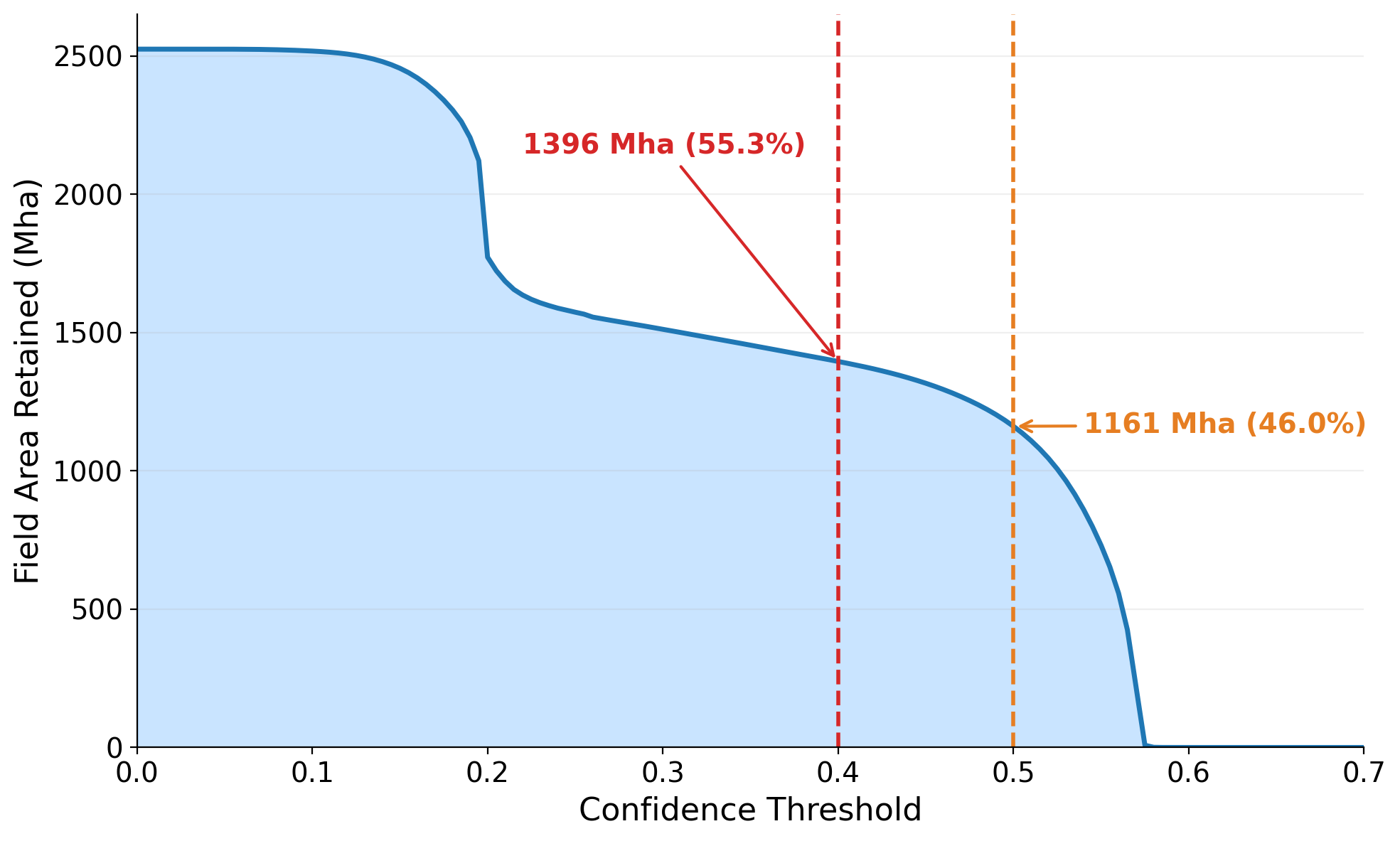}
  \caption{\textbf{Cumulative retention as a function of confidence threshold for the 2025 predictions.}
    (Left) Number of fields retained when discarding all predictions below a given confidence threshold:
    at a threshold of 0.40, 0.84\,B fields (54.3\% of the total 1.55\,B) are retained;
    at 0.50, 0.35\,B fields (22.6\%) remain.
    (Right) Total mapped field area retained under the same thresholds:
    at 0.40, 1\,396\,Mha (55.3\% of the total) are retained.
    The steep drop-off between 0.40 and 0.55 in both curves indicates that a large fraction of predictions cluster near the decision boundary; the near-identical retention fractions for fields and area at the same threshold indicate that filtering does not preferentially remove small-area polygons.}
  \label{fig:confidence-cdf}
\end{figure}

All released layers share the common 500\,m global grid defined above and are available as Cloud-Optimized GeoTIFFs (COGs) with internal tiling and overviews for efficient web access.
We released the quality indicator layers described above (field prediction density, model entropy, cropland consensus agreement, and crop consensus count), together with a confidence score layer and confidence-filtered density layers.
We produced the confidence score layer by applying the Random Forest model (model-only features, crop $\leq 2$ filter; Table~\ref{tab:confidence_model}) to the seven model-derived quality indicators globally.
The confidence-filtered density layers mirror the field prediction density layer but zero out cells where the confidence score falls below a threshold (0.4 and 0.5 variants available); the 0.5 variant removes approximately 25\% of active cells, providing a more conservative field area estimate.
Figure~\ref{fig:confidence-cdf} shows the cumulative distributions of predicted field counts and total mapped field area as functions of the confidence threshold, illustrating the sensitivity of the released filtered products to the chosen cutoff.

\subsection*{Validation analysis}

We rasterized ground-truth (GT) field boundary polygons from 24 countries onto the 500\,m grid, marking any cell that overlaps a polygon boundary.
To define spatial coverage regions, we clustered GT polygons per country using DBSCAN ($\varepsilon = 0.1^{\circ} \approx 10$\,km, min\_samples $= 3$) and computed a buffered convex hull (buffer $= 0.025^{\circ} \approx 2.5$\,km) per cluster.
We restricted the analysis to cells within these coverage hulls where the model predicted field content (field pixel count $> 0$), and labeled each cell as \textit{field} (any GT polygon touches the cell) or \textit{non-field} (model active within hull, no GT overlap).

To train the confidence models, we used balanced subsampling (up to 5,000 per class per country), yielding training sets of $\sim$180,000 cells.
We evaluated all four feature configurations (Table~\ref{tab:feature_sets}) with logistic regression and Random Forest (200 trees, max depth 10, min samples per leaf 20), using 5-fold stratified cross-validation and leave-one-country-out (LOCO) cross-validation.

\paragraph{Training data construction for confidence model.}
Positive training examples are 500\,m cells that overlap at least one GT polygon.
Constructing reliable negative examples is more challenging: cells within the coverage hull but outside GT polygons are not necessarily non-field, since ground-truth coverage is often incomplete~\citep{olofsson2014good}.
To obtain high-confidence true negatives, we applied a consensus-based filter using the crop consensus count layer, retaining non-field cells only if their mean crop count was $\leq 2$ and removing cells where the independent cropland consensus strongly indicates cropland.
This filter retained 58,521 non-field cells (10.6\% of the original 553,313) across all 24 countries.
We also tested filter thresholds of $\leq 3$ and $\leq 1$ (Table~\ref{tab:confidence_full}).

\paragraph{Confidence model results.}
Table~\ref{tab:confidence_full} reports confidence model performance across all feature configurations and filter thresholds.
Without filtering, all configurations achieved AUC in the range 0.56--0.61, reflecting noise in the non-field training labels.
Performance improved substantially with filtering: at crop $\leq 2$ with Model-only features, the Random Forest achieved AUC = 0.822; with the All feature set, AUC reached 0.964.

\begin{table}[!ht]
\centering
\caption{\textbf{Full confidence model results} across all filter thresholds and feature configurations (5-fold cross-validation).}
\label{tab:confidence_full}
\scriptsize
\begin{tabular}{lllcccc}
\toprule
\textbf{Filter} & \textbf{Features} & \textbf{Model} & \textbf{AUC} & \textbf{F1} & \textbf{Precision} & \textbf{Recall} \\
\midrule
  Unfiltered & Model+P/R & Logistic Regression & 0.58$\pm$0.01 & 0.64$\pm$0.00 & 0.55$\pm$0.00 & 0.76$\pm$0.01 \\
   & Model+P/R & Random Forest & 0.60$\pm$0.00 & 0.63$\pm$0.00 & 0.55$\pm$0.00 & 0.74$\pm$0.01 \\
   & All & Logistic Regression & 0.59$\pm$0.01 & 0.63$\pm$0.00 & 0.55$\pm$0.00 & 0.75$\pm$0.00 \\
   & All & Random Forest & 0.61$\pm$0.00 & 0.64$\pm$0.00 & 0.56$\pm$0.00 & 0.75$\pm$0.01 \\
   & Model+consensus & Logistic Regression & 0.59$\pm$0.00 & 0.61$\pm$0.00 & 0.55$\pm$0.00 & 0.67$\pm$0.01 \\
   & Model+consensus & Random Forest & 0.60$\pm$0.00 & 0.63$\pm$0.01 & 0.56$\pm$0.00 & 0.73$\pm$0.01 \\
   & Model-only & Logistic Regression & 0.56$\pm$0.01 & 0.59$\pm$0.00 & 0.54$\pm$0.00 & 0.64$\pm$0.01 \\
   & Model-only & Random Forest & 0.58$\pm$0.00 & 0.60$\pm$0.00 & 0.55$\pm$0.00 & 0.67$\pm$0.00 \\
\addlinespace
  Crop $\leq 3$ & Model+P/R & Logistic Regression & 0.90$\pm$0.00 & 0.82$\pm$0.00 & 0.84$\pm$0.00 & 0.81$\pm$0.00 \\
   & Model+P/R & Random Forest & 0.92$\pm$0.00 & 0.85$\pm$0.00 & 0.94$\pm$0.00 & 0.77$\pm$0.01 \\
   & All & Logistic Regression & 0.91$\pm$0.00 & 0.85$\pm$0.00 & 0.92$\pm$0.00 & 0.80$\pm$0.00 \\
   & All & Random Forest & 0.93$\pm$0.00 & 0.88$\pm$0.00 & 0.99$\pm$0.00 & 0.79$\pm$0.01 \\
   & Model+consensus & Logistic Regression & 0.91$\pm$0.00 & 0.85$\pm$0.00 & 0.91$\pm$0.00 & 0.80$\pm$0.00 \\
   & Model+consensus & Random Forest & 0.93$\pm$0.00 & 0.87$\pm$0.00 & 0.99$\pm$0.00 & 0.78$\pm$0.01 \\
   & Model-only & Logistic Regression & 0.76$\pm$0.00 & 0.71$\pm$0.00 & 0.71$\pm$0.00 & 0.70$\pm$0.01 \\
   & Model-only & Random Forest & 0.80$\pm$0.00 & 0.74$\pm$0.00 & 0.74$\pm$0.01 & 0.74$\pm$0.01 \\
\addlinespace
  Crop $\leq 2$ & Model+P/R & Logistic Regression & 0.94$\pm$0.00 & 0.87$\pm$0.00 & 0.86$\pm$0.00 & 0.89$\pm$0.00 \\
   & Model+P/R & Random Forest & 0.96$\pm$0.00 & 0.91$\pm$0.00 & 0.98$\pm$0.00 & 0.86$\pm$0.00 \\
   & All & Logistic Regression & 0.95$\pm$0.00 & 0.92$\pm$0.00 & 0.98$\pm$0.00 & 0.87$\pm$0.00 \\
   & All & Random Forest & 0.96$\pm$0.00 & 0.93$\pm$0.00 & 1.00$\pm$0.00 & 0.87$\pm$0.00 \\
   & Model+consensus & Logistic Regression & 0.95$\pm$0.00 & 0.92$\pm$0.00 & 0.99$\pm$0.00 & 0.86$\pm$0.00 \\
   & Model+consensus & Random Forest & 0.96$\pm$0.00 & 0.93$\pm$0.00 & 1.00$\pm$0.00 & 0.87$\pm$0.00 \\
   & Model-only & Logistic Regression & 0.75$\pm$0.00 & 0.71$\pm$0.01 & 0.70$\pm$0.00 & 0.71$\pm$0.01 \\
   & Model-only & Random Forest & 0.82$\pm$0.00 & 0.76$\pm$0.00 & 0.74$\pm$0.00 & 0.78$\pm$0.01 \\
\addlinespace
  Crop $\leq 1$ & Model+P/R & Logistic Regression & 0.97$\pm$0.00 & 0.93$\pm$0.00 & 0.94$\pm$0.01 & 0.93$\pm$0.00 \\
   & Model+P/R & Random Forest & 0.98$\pm$0.00 & 0.96$\pm$0.00 & 0.99$\pm$0.00 & 0.93$\pm$0.00 \\
   & All & Logistic Regression & 0.98$\pm$0.00 & 0.97$\pm$0.00 & 1.00$\pm$0.00 & 0.94$\pm$0.00 \\
   & All & Random Forest & 0.99$\pm$0.00 & 0.98$\pm$0.00 & 1.00$\pm$0.00 & 0.96$\pm$0.00 \\
   & Model+consensus & Logistic Regression & 0.98$\pm$0.00 & 0.97$\pm$0.00 & 1.00$\pm$0.00 & 0.94$\pm$0.00 \\
   & Model+consensus & Random Forest & 0.98$\pm$0.00 & 0.98$\pm$0.00 & 1.00$\pm$0.00 & 0.96$\pm$0.00 \\
   & Model-only & Logistic Regression & 0.74$\pm$0.01 & 0.71$\pm$0.01 & 0.69$\pm$0.01 & 0.73$\pm$0.01 \\
   & Model-only & Random Forest & 0.85$\pm$0.01 & 0.76$\pm$0.00 & 0.78$\pm$0.01 & 0.75$\pm$0.01 \\
\addlinespace
\bottomrule
\end{tabular}
\end{table}

\paragraph{Leave-one-country-out evaluation.}
LOCO cross-validation for the recommended configuration (crop $\leq 2$ filter, model-only features, Random Forest) yielded mean AUC = 0.842, with per-country AUC ranging from 0.64 (Brazil) to 0.95 (Slovakia).
Including external consensus features (All feature set) raised the LOCO mean AUC to 0.958.
Portugal was excluded from LOCO evaluation because its filtered training set contains only 6 field samples and no non-field samples, making per-country AUC undefined; it was also excluded from the per-country recall analysis because its ground-truth coverage (818\,K pixels from 5{,}040 polygons in the Azores) yielded zero recall under the 2025 predictions.

\subsection*{Full-country pixel-level evaluation}

To complement the hull-restricted per-country recall analysis with a full-country precision and recall assessment, we evaluated the PRUE 2024 predictions against complete national Land Parcel Identification System (LPIS) or INVEKOS databases for Austria, Finland, and Latvia.
These three countries were selected because their 2024 national agricultural parcel databases are publicly available, each parcel is associated with a machine-readable crop type, and together the three countries span a range of agro-climatic conditions from temperate lowlands to the boreal zone.

\paragraph{Ground truth filtering.}
FTW's training data covers only annual crops (examples include wheat, rice, maize, soybeans, and barley) and excludes permanent and perennial crops such as fruit and nut trees, as well as pasture, grazing, fallow, orchards, vineyards, and forestry~\citep{kerner2025fields}.
To match this training scope, we filtered each national parcel database to retain only seasonal crops: 690,022 of 2,956,449 parcels in Austria (INVEKOS), 213,097 of 420,863 in Latvia (LPIS), and 460,337 of 1,086,825 in Finland (LPIS).
Under this design, a PRUE-predicted field pixel that falls on an excluded parcel---for example, permanent grassland, orchards, tree nurseries, perennial forage, or landscape elements---was counted as a false positive.

\paragraph{Rasterization and masking.}
We reprojected each national parcel database from its native CRS (EPSG:31287 for Austria, EPSG:3067 for Finland, EPSG:3059 for Latvia) to EPSG:4326 and rasterized the filtered polygons onto the ESA WorldCover 10\,m grid (using the \textit{all touched} rule).
We rasterized national boundaries to the same grid to define the evaluation mask; pixels outside the national boundary were excluded so that predictions spilling into neighboring countries that share the same WorldCover tile were not counted.
We treated PRUE classes 1 (field interior) and 2 (field boundary) as positive and class 0 (background) as negative, and reported pixel-level precision, recall, F1, and intersection-over-union (IoU) at the native 10\,m resolution.

\paragraph{Confidence filtering.}
We read the 500\,m confidence raster for each tile extent, upsampled to 10\,m by nearest-neighbor repetition, and retained only predictions where the confidence exceeded the threshold (conf\,$\geq$\,0.4 in the main results; additional thresholds 0.3, 0.5, and 0.55 were computed per tile for sensitivity analysis).
Country-wide metrics (Table~\ref{tab:full_country_metrics}) are the sum-pooled pixel confusion matrices across all tiles.

\subsection*{Distributional comparison in Zambia}

Public ground-truth field boundaries for Zambia are not available.
To characterize PRUE's behavior in a smallholder Sub-Saharan context, we compared the PRUE 2025 Zambia polygons (39.4\,million field polygons before filtering; 6.9\,million at conf\,$\geq$\,0.4) to an independently produced 2024 ML-derived field boundary dataset (7.7\,million polygons) covering the same national extent.
Country assignment for PRUE polygons used centroid-based lookup against an ADM0 raster at 250\,m.
We reprojected a random sample of 500\,000 PRUE polygons to UTM zone 35S (EPSG:32735) to compute metric-unit area, perimeter, Polsby--Popper compactness~\citep{polsby1991third}---a $4\pi A/P^2$ shape index that equals 1 for a perfect circle and approaches 0 for long or highly irregular shapes---shape index ($P/2\sqrt{\pi A}$), and boundary fractal dimension; the reference dataset provides these attributes natively.
The comparison is descriptive: neither product is ground truth, and systematic differences reflect differences in training data, year, and modeling approach rather than absolute accuracy.

\end{document}